\documentclass{article}

\usepackage{graphicx}
\usepackage[width=.75\textwidth,font=small]{caption}                                 
\usepackage{amsmath, amsthm, amsfonts, soul}
\usepackage{titling}
\usepackage{url}
\usepackage{float}
\usepackage{hyperref}
\usepackage[utf8]{inputenc}
\usepackage[T1]{fontenc}
\usepackage[english]{babel}
\usepackage{indentfirst}
\usepackage[usenames,dvipsnames]{color}
\usepackage{enumitem}
\usepackage{amsfonts}
\usepackage{amsmath}
\usepackage{multicol}
\usepackage{algpseudocode}
\usepackage{csvsimple}
\usepackage{fancyvrb}
\usepackage{lipsum}
\usepackage{listings}

\theoremstyle{definition}

\theoremstyle{remark}

\pretitle{%
  \begin{center}
  \LARGE
  \includegraphics[height=3cm]{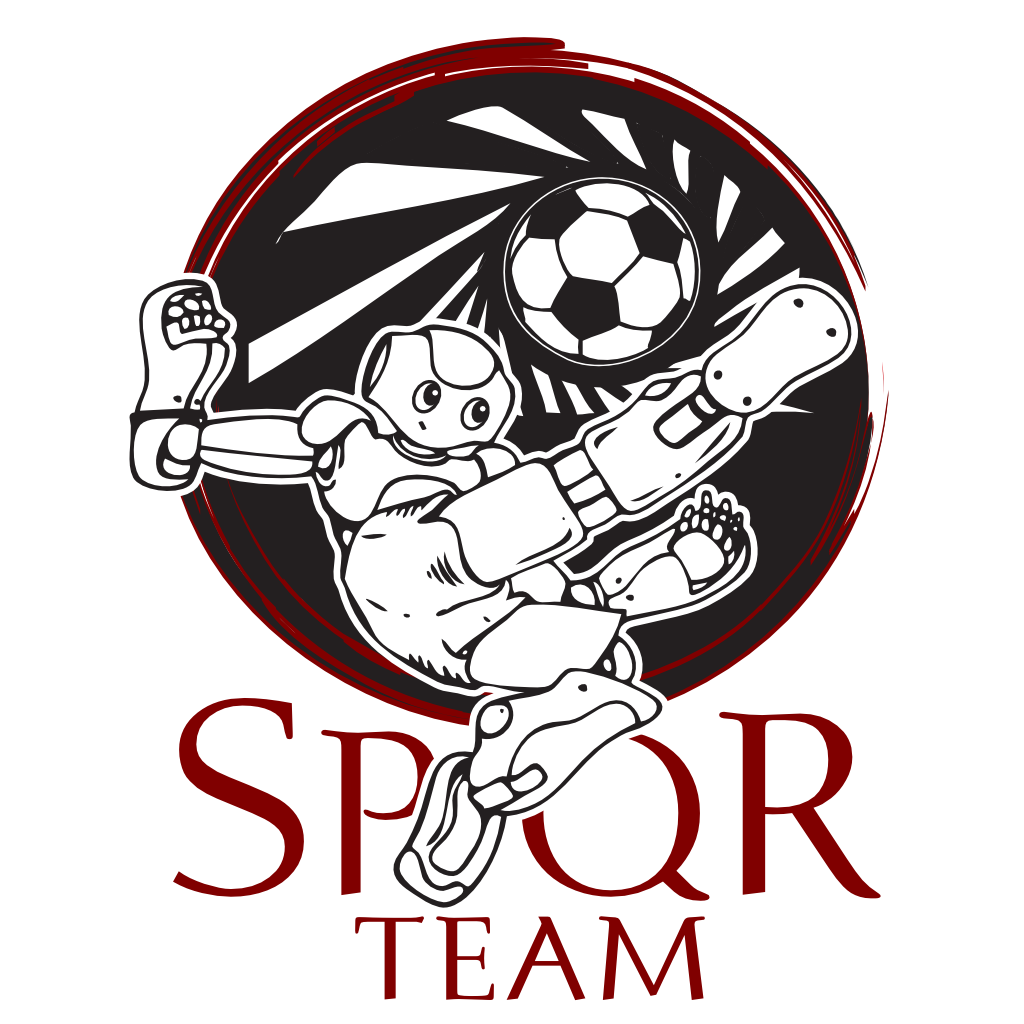}\\[\bigskipamount]
  \includegraphics[height=3cm]{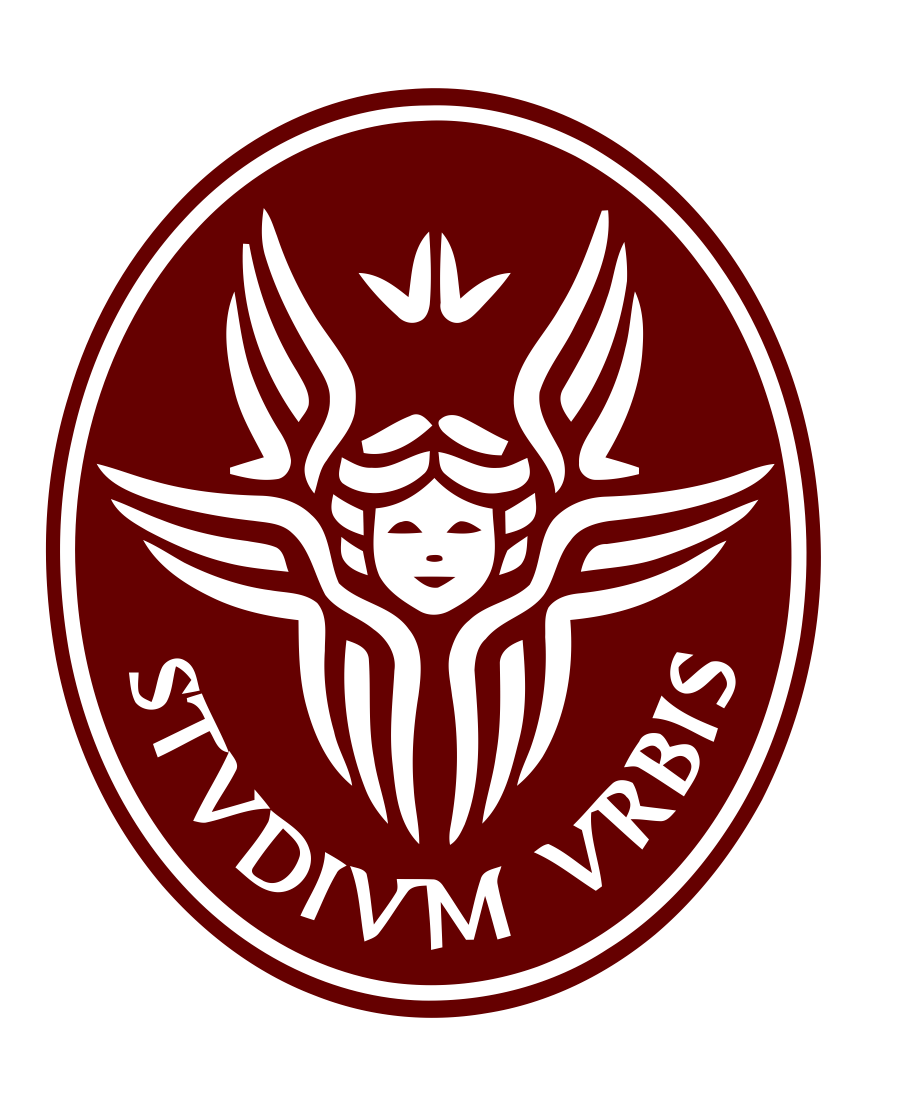}\\[\bigskipamount]
}

\begin{document}

\begin{table}[h!]
  \centering 
  \begin{tabular}{ c c c}
    \begin{minipage}{.15\textwidth}
      \includegraphics[width=1.1\textwidth]{sapienza}
    \end{minipage}
    &
    \begin{minipage}{7.3cm}
              \centering{
              \large\textbf{SPQR RoboCup SPL Team\\  }
              \vspace{8 mm}
              \Large \textbf{Machine Learning for Realistic Ball Detection in RoboCup SPL} \\
              \vspace{5 mm}
		}
		
	 Domenico Bloisi, Francesco Del Duchetto, 
	 Tiziano Manoni, Vincenzo Suriani
		\\[0.3cm]
		  \begin{normalsize}
		  \textit{Department of Computer, Control and Management Engineering ``Antonio Ruberti''\\
		    Sapienza University of Rome\\
			  Via Ariosto 25, 00185 Rome, Italy}\\
			  \textbf{website:} \url{http://spqr.diag.uniroma1.it}
			  \end{normalsize}
    \end{minipage}
    & 
    \begin{minipage}{.15\textwidth}
       \includegraphics[width=1.1\textwidth]{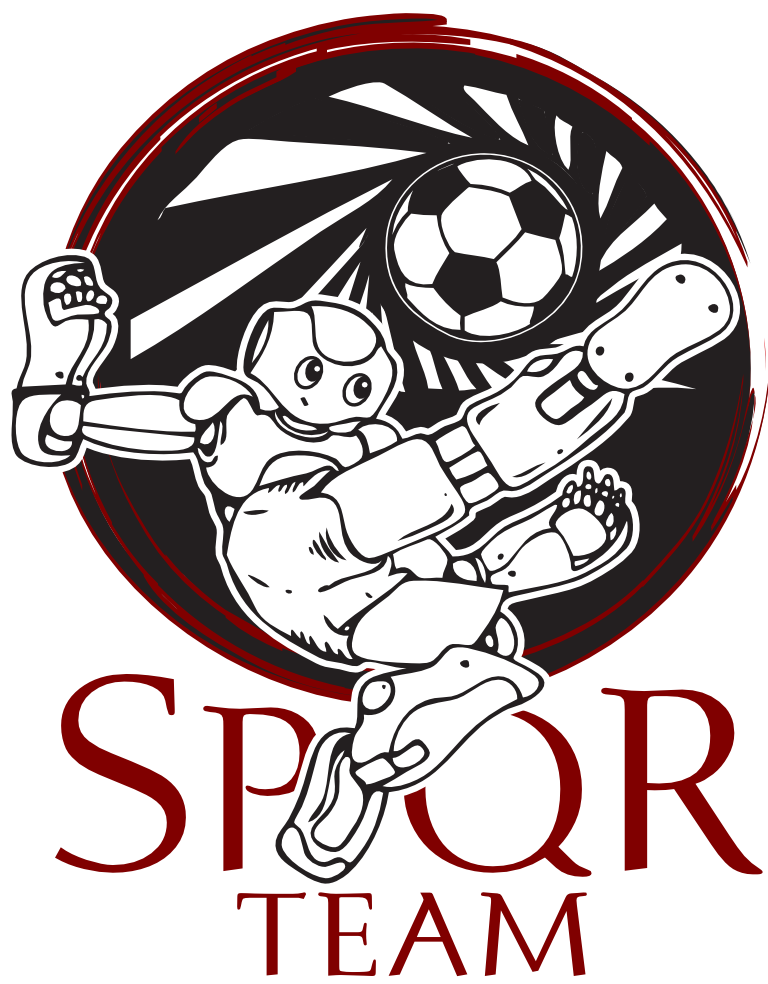}
    \end{minipage}
  \end{tabular}
\end{table}

\abstract{
In this technical report, we describe the use of a machine learning approach for
detecting the realistic black and white ball currently in use in the RoboCup
Standard Platform League. Our aim is to provide a ready-to-use software module
that can be useful for the RoboCup SPL community. To this end,
the approach is integrated within the official B-Human code release 2016.
The complete code for the approach
presented in this work can be downloaded from the SPQR Team homepage at
\url{http://spqr.diag.uniroma1.it} and from the SPQR Team GitHub repository
at \url{https://github.com/SPQRTeam/SPQRBallPerceptor}.
The approach has been tested in multiple environments, both indoor and outdoor.
Furthermore, the ball detector described in this technical report has been used
by the SPQR Robot Soccer Team during the competitions of the Robocup German Open 2017.
To facilitate the use of our code by other teams, we have prepared a
step-by-step installation guide.}
\newpage

\tableofcontents
\clearpage

\pagenumbering{arabic}

\section{Introduction}
\begin{figure}[h!]
\centering
\includegraphics[width=0.8\linewidth]{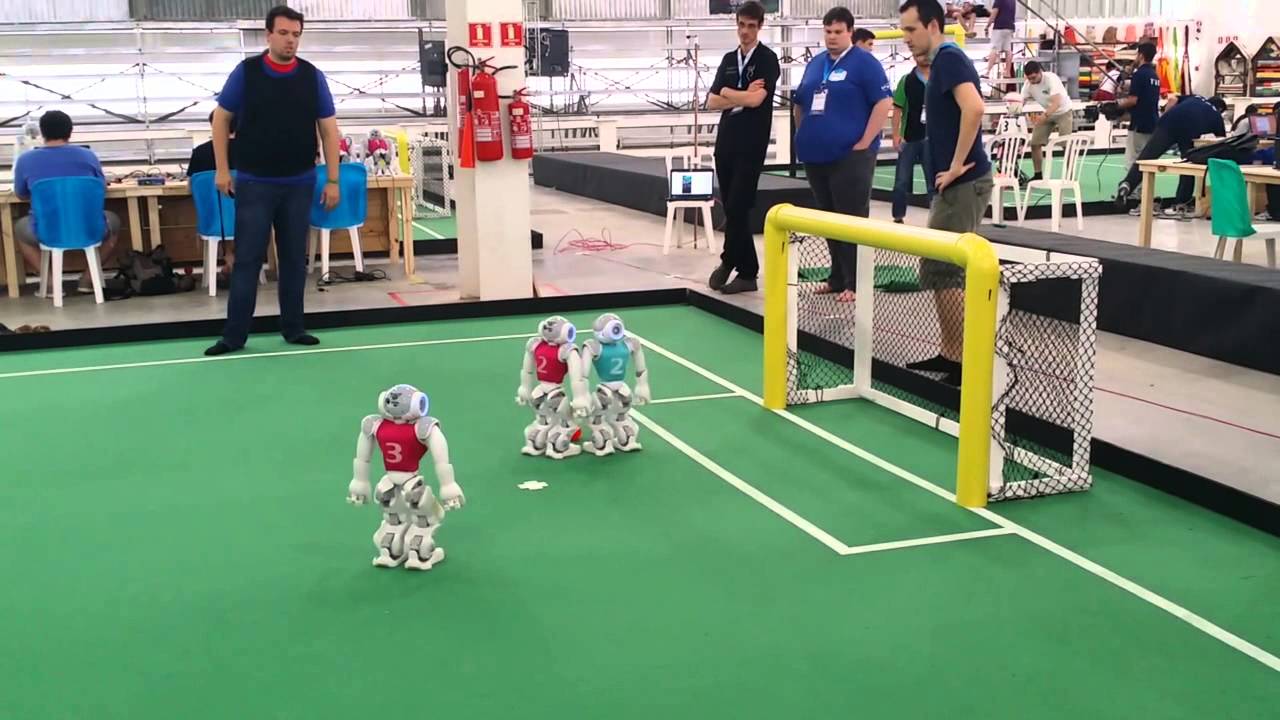}
\caption{The former field set up for the SPL: red ball, controlled illumination, and yellow colored goals.
This image from the SPQR Team - Qualification Video 2016}
\label{fig:redball}
\end{figure}

Starting from 2008, the RoboCup Standard Platform League (SPL) involves
Aldebaran NAO as the common robot for the competitions. Due to the limited
computational power available, Computer Vision techniques adopted by the different teams are mostly based on color segmentation approaches. The success
of those solutions has been facilitated by special expedients, for example the
use of a red ball, controlled illumination, and yellow colored goals (see Fig. \ref{fig:redball}).

\begin{figure}[h!]
\centering
\includegraphics[width=0.8\linewidth]{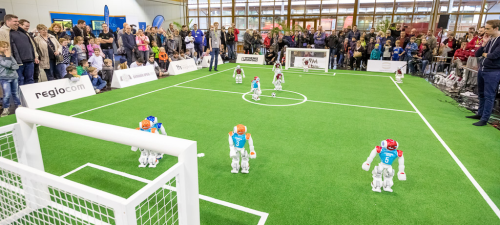}
\caption{The current game fields includes white goal posts.
This image from the RoboCup German Open 2017 website}
\label{fig:whiteposts}
\end{figure}

\begin{figure}[h!]
\centering
\includegraphics[width=0.8\linewidth]{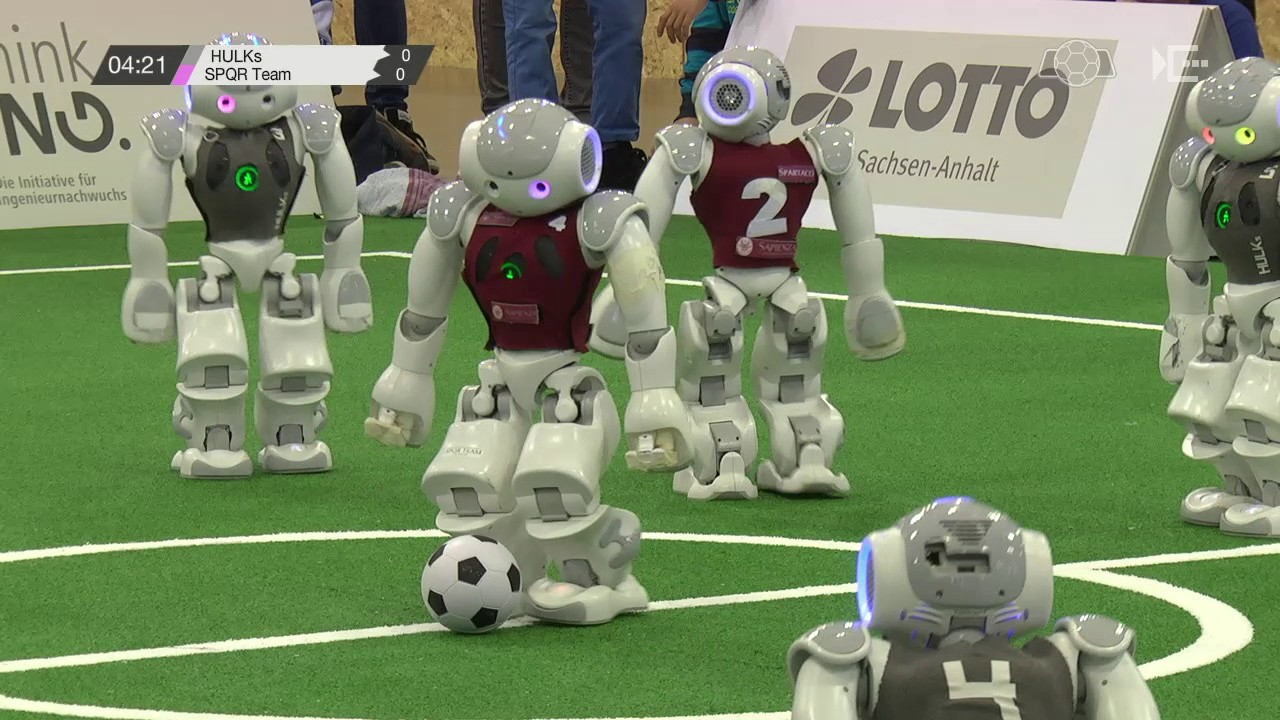}
\caption{A ball with black and white patches and personalized jersey shirts are used during the game.
This image from the Streamteam HTWK YouTube channel}
\label{fig:jersey}
\end{figure}

However,
the current trend in using more and more realistic game fields, with white goal
posts (see Fig. \ref{fig:whiteposts}), natural light, and a ball with black and white patches, as well as personalized jersey shirts (see Fig. \ref{fig:jersey}), imposes the adoption of robust detection and classification
methods, which can deal with a more realistic and complex environment.

In this technical report, we propose a supervised method for detecting
the realistic black and white ball in images captured by a NAO robot.
Our approach is designed to work in scenes with changes in the lighting conditions that can
affect significantly the robot perception.
To train and test our method, we have created a
specific data set, containing images coming from the upper
and lower cameras of different NAO robots in action.
The data set is fully annotated and publicly
available as part of the SPQR NAO image data set\footnote{\url{http://www.dis.uniroma1.it/~labrococo/?q=node/459}}.
It is worth to notice that our approach works with different light conditions 
without color calibrations and camera parameters settings.

The rest of the report is organized as follows.
Section \ref{sec:relatedwork} presents an overview
of the object detection methods presented during the last years of Robocup
competition. Section \ref{sec:method} contains the details of our method.
The training stage is described in Section \ref{sec:training}, while
experimental tests are discussed in Section \ref{sec:experiments}.
A step-by-step guide to install the software is
given in Section \ref{sec:code}. 
Conclusions are drawn in Section \ref{sec:conclusions}.

\section{Background and Related Work}
\label{sec:relatedwork}

SPQR Team contributed with the development of a dynamic color segmentation method \cite{iocchi2006robust} that is able to provide robust and efficient color segmentation with very little calibration effort, as well as a hierarchical approach for examining the image pixels using a set of sentinel pixels that are non-uniformly spread on the image.

Since 2013, we have adopted the B-Human framework.
In the same year we presented an open source software for monitoring humanoid
soccer robot behaviours called GNAO (Ground Truth Acquisition System for NAO Soccer Robots) \cite{Pennisi2014}.
The software is designed for evaluating and testing robot coordination
and localization. The hardware architecture
consists of multiple low-cost visual sensors (four Kinects). The
software includes a foreground computation module and a detection unit
for both players and ball. A graphical user interface has been developed
in order to facilitate the creation of a shared multi-camera plan view,
in which the observations of players and ball are re-projected to obtain
global positions.

In 2016, we contribute by presenting in \cite{albani2016deep} a novel approach for object detection and classification based on Convolutional Neural Networks (CNN), combined with an image region segmentation for reducing the search space. Furthermore, the used data set of annotated images captured in real conditions from NAO robots in action was made available for the community.

This work is based on the B-Human code release 2016\footnote{https://github.com/bhuman/BHumanCodeRelease}
and has been tested and validated on Ubuntu 16.04 LTS.
We wrote a module running within the B-Human Cognition process that provides the Ball position and velocity.
Our module uses the available primitives within the B-Human
framework plus some OpenCV version 2.4 functions.
In order to use OpenCV within the B-Human framework we have
accessed the OpenCV library that is shipped with the NAOqi-SDK.

\subsection{OpenCV}
OpenCV\footnote{http://opencv.org/} (Open Source Computer Vision) is a library of programming functions mainly aimed at real-time
computer vision. The library is cross-platform and free for use under the open-source BSD license.
It was officially launched in 1999 by Intel Research, with the aim of advancing CPU-intensive applications.
The goals of the project are  described as:
\begin{itemize}
  \item Advancing vision research by providing not only open but also optimized code.
  \item Providing a common infrastructure that developers could build 
      on, so that code would be more readily readable and transferable.
  \item Advance vision-based commercial applications by making portable, 
      performance-optimized code available for free.
\end{itemize}
The library at present time covers most of the computer vision algorithms in the literature ranging from
facial/gesture recognition, segmentation, motion tracking to artificial neural networks, decision trees, 
KNN, SVM and others.

\section{Method}
\label{sec:method}

In order to speed up the ball detection process,
the size of the camera frames are lowered to $320\times 240$ and
$160\times 120$ for the upper and lower camera, respectively.
The images captured from the NAO's cameras are converted in grayscale.

\subsection{Cascade}

The first step is to train our classifier with a few thousand sample views of the ball, called positive examples, with a particular size, and also negative examples (i.e., images without a ball).

After this training, a classifier can be applied to a particular region in an input image. In our case, we applied it to each frame of our camera.
The classifier outputs can be a "1" or a "0": if the region is likely to show a ball, we will assign a "1",
otherwise we will have a "0".

Generally, the size of a frame is bigger then images using to train our classifier, so in order to be able to find the objects of interest at different sizes and a different coordinate in the frame, this method search for the object in the whole image by moving the search window across the image. To find an object of an unknown size in the image the scan procedure should be done several times at different scales.

The word ``cascade'' in the classifier name means that the resultant classifier consists of several simpler classifiers (stages) that are applied subsequently to a region of interest until at some stage the candidate is rejected or all the stages are passed.

We partially rewrote the OpenCV classifier in order to remove some function or dependencies that we do not use and make everything faster and less computationally demanding. 

In our Ball Detector we use a Local Binary Patterns (LBP) classifier, which is a type of visual descriptor used for classification in Computer Vision.
With respect to a Haar feature based classifier, LBP is faster, but less accurate.

\subsection{Patches}
Searching the ball inside the image is a very expensive task if we have to check every possible position,
and if for each position we look for all possible sizes. The \texttt{BallSpotPatchesProvider} module 
provides a list of patches in which is probable to find a ball in the current frame. Each patch is
encapsulated into an \texttt{ImagePatch} instance which provides the position and size of the patch 
along the minimum and maximum ball sizes that can be present inside it, the neighbors threshold for 
the cascade and the type of patch (i.e. which predictor produced the patch). The neighbors values are
chosen according to the type of patch and the presence of obstacles inside the patch. A large min neighbors
value increase the accuracy (less false positives) while slowing down the perceptor module. Therefore, the
value is increased when the frame contains obstacles, where is probable to have false balls, and otherwise 
kept low. 

The patches are inserted in the representation in a specific order: from the fastest to check to the
most expensive one. The ball perceptor module should be passed on each patch in the same order they arrive 
and stop as soon as it finds a ball. This mechanism allows to reliably track the ball during a match while
keeping the computation as low as possible. The prediction on how fast is to check the patch is mostly based
on the size of the patch.

The ball perceptor module is able to detect small balls with up to 16 pixels size. We normally use the
low resolution images ($320\times 240$ and $160\times 120$) for performances issues, but on small patches we still can 
use the high resolution images in order to detect balls up to half of the size with respect to the 
low resolution images.   

\subsubsection{Patches predictors}
The decision on where in the image we should take a patch is assigned to some methods which provides 
a prediction on where the ball could be. We found the predictors listed below sufficient for finding 
and tracking the ball.
\begin{itemize}
  \item \textbf{lastFramePrediction \quad} If in the last two frames we detected a ball we simply 
      predict the ball to be in the last seen position. When present it is the first patch in the 
      \texttt{ImagePatches} list. This very simple heuristic it's very useful for
      building continuity in the perception of the ball between consecutive frames.
  \item \textbf{kalmanPrediction \quad} We take advantage of the \texttt{TeamBallModel} representation
      provided by the \textsc{B-Human 2016} framework for looking at the position in the field were
      the consensus about the ball position between the robots is.
  \item \textbf{framePatrolling \quad} The patrolling patch is the last one present in the 
      \texttt{ImagePatches} list, because it is the most expensive to check and our last chance
      to find a ball in the current frame. Given that controlling the entire frame it too costly, the idea
      is to cover the whole field of view in a small number $n$ of consecutive frames. Therefore we 
      subdivide the frame in $n$ overlapping patches and at each time step we check a different one. As 
      soon as we lose a ball we restart with the patroling patch which contains positions on the field
      closer to the robot.
\end{itemize}

\section{Training}
\label{sec:training}
This section provides the details needed to train a classifier from scratch.
The code released in the SPQR Ball Perceptor repository already
contains the pre-trained models for the two cameras.

The pre-trained classifier can be also downloaded from the
the SPQR NAO image data set at \url{http://www.dis.uniroma1.it/~labrococo/?q=node/459}.
The models are in the standard OpenCV \texttt{xml} format and can be found inside the \texttt{Config} 
folder.

In order to train the classifier, we used an approach based on OpenCV for the cascade classifier training.
The details about the OpenCV approach can be found at \url{http://docs.opencv.org/2.4.13.2/doc/user_guide/ug_traincascade.html}.
The following brief tutorial shows how to train again the model for the Ball Perceptor.
An html version of the training procedure can be found at
\url{http://profs.scienze.univr.it/~bloisi/tutorial/balldetection.html}

We assume the following directory structure:\\
$/balldetection\\
\indent    pos\_top\\
\indent\indent        img1.png\\
\indent\indent        img2.png\\
\indent\indent        ...\\
\indent    neg\_top\\
\indent\indent        img3.png\\
\indent\indent        img4.png\\
\indent\indent        ...\\
\indent    createpos.cpp$

\subsection{Create the positive training data}

All the image patches in the set of positive samples have to be listed in a text file. This text file should be formatted to adere to the needs of the OpenCV function \textit{opencv\_createsamples}.
You can use the C++ file \textit{createpos.cpp} to create a file named "positives.txt" in a format suitable for OpenCV.
Use for example
\begin{verbatim}
$ g++ createpos.cpp -o createpos -lws2_32
\end{verbatim}
for compiling it under Windows with MinGW.\\
The command lists for creating the text file is:
\begin{verbatim}
$ cd balldetection

$ createpos pos_top

$ opencv_createsamples -info positives.txt -w 16 -h 16 -vec pos.vec -num 2728
\end{verbatim}
where
\begin{itemize}
\item \textbf{positives.txt} contains the list of the positive patches that will be used for training the classifier
\item \textbf{w} is the width of the search window
\item \textbf{h} is the height of the search window
\item \textbf{vec} is a vector file of positive samples in the OpenCV format
\item \textbf{num} is the total number of positive samples
\end{itemize} 

We experimentally found that 16$\times$16 is a good window size for our application scenario where we want to find the black and white ball.
Once completed the above steps, our directory structure is:
$/balldetection
\indent    pos\_top\\
\indent    neg\_top\\
\indent    createpos.cpp\\
\indent    positives.txt\\
\indent    pos.vec\\ $

\subsection{Create the negative training data}
All the images in the set of negative samples have to be listed in a text file.
For example, in Windows the command lists for creating the text file is:
\begin{verbatim}
$ dir /b/s .\neg_top \*.png > negatives.txt
\end{verbatim}
Once completed the above steps, our directory structure is:\\
$/balldetection\\
\indent    pos\_top\\
\indent    neg\_top\\
\indent    createpos.cpp\\
\indent    positives.txt\\
\indent    pos.vec\\
\indent    negatives.txt$

\subsection{Train the classifier}
To train the classifier it is possible to use the OpenCV method \textit{opencv\_traincascade}, but before we need to create an empty folder named "classifier".
Our directory structure is now:\\
$/balldetection\\
\indent    classifier\\
\indent    pos\_top\\
\indent    neg\_top\\
\indent    createpos.cpp\\
\indent    positives.txt\\
\indent    pos.vec\\
\indent    negatives.txt$
\begin{verbatim}
$ opencv_traincascade -data classifier -vec pos.vec -bg \
negatives.txt -numStages 20 -minHitRate 0.999 \
-maxFalseAlarmRate 0.5 -numPos 2728 -numNeg 16000 -w 16 \
-h 16 -mode ALL -precalcValBufSize 256 -precalcIdxBufSize 256 \ 
-acceptanceRatioBreakValue 10e-5 -nonsym -baseFormatSave \
-featureType LBP
\end{verbatim}

Please refer to the OpenCV documentation for the specific meaning of the parameters.
It is important to note the use of the parameter nonsym
since in the case of the black and white ball the pattern is not symmetric in all possible views.

\section{Experimental Results}
\label{sec:experiments}

The SPQR Ball Perceptor has been used by the SPQR Team during the competitions of the Robocup German Open 2017.
Before the RoboCup competitions, it has been widely tested in indoor and outdoor scenarios.
The SPQR Ball Perceptor does not require color calibration to detect the ball. Developed to deal with the league trend to move competitions towards natural lighting conditions, it appears suitable to be used in matches characterized
by consistent variations in illumination.

\begin{figure}[h!]
\centering
\frame{\includegraphics[width=0.8\linewidth]{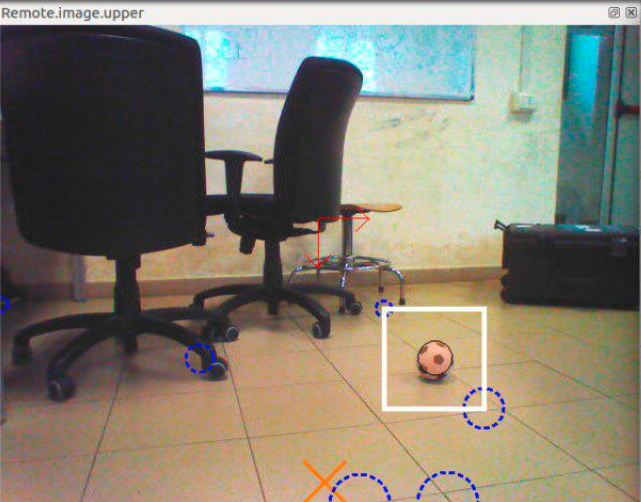}}
\caption{Ball perceived with upper camera in our lab. No color calibration is needed. The camera matrix has to be calibrated.}
\end{figure}

\begin{figure}[h!]
\centering
\frame{\includegraphics[width=0.8\linewidth]{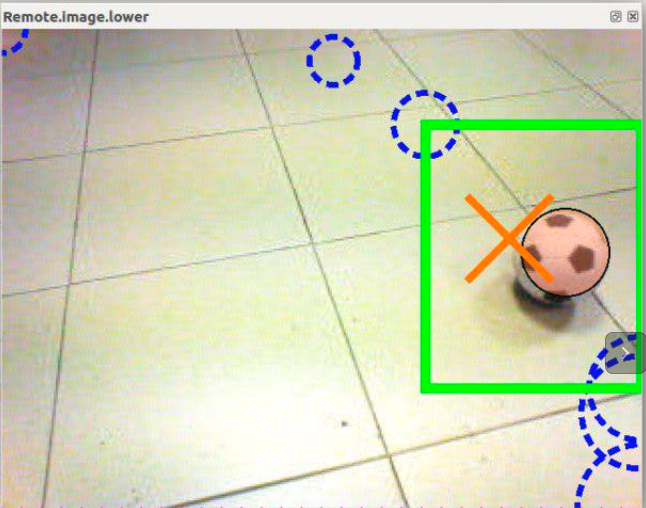}}
\caption{Ball perceived with lower camera in our lab.}
\end{figure}

\begin{figure}[h!]
\centering
\frame{\includegraphics[width=1.0\linewidth]{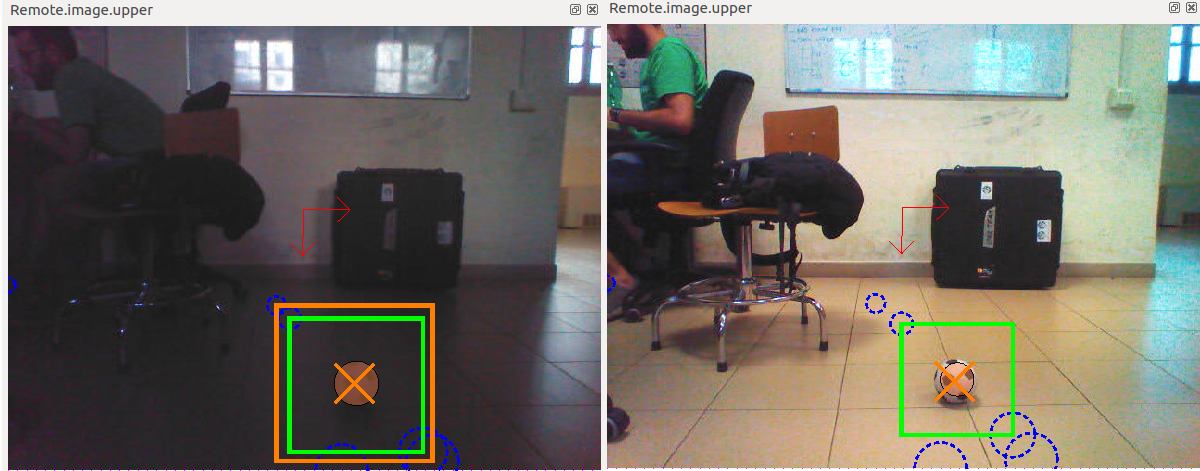}}
\caption{Test with different light conditions. The ball is detected without changes in the camera settings both
in a very bright environment and in the same environment with darker light conditions.}
\end{figure}

An indoor/outdoor use of the \textit{SPQRBallPerceptor} can be seen at \url{https://www.youtube.com/watch?v=fIgEwHRe6Bk}.
A video with the SPQR Team highlights during the RoboCup German Open 2017
is available at\\ \url{https://www.youtube.com/watch?v=KFxiFpezvw0}

\newpage
\section{Code and Installation}
\label{sec:code}

The SPQR Ball Perceptor has been tested on Ubuntu 16.04 LTS. The following is the installation procedure.\\
\begin{itemize}
\item The following dependencies are required:
\begin{itemize}
\item cmake
\item libgtk2.0-dev 
\item pkg-config 
\item libavcodec-dev 
\item libavformat-dev 
\item libswscale-dev 
\item libjpeg8 
\item libjpeg-turbo8-dbg 
\item lib32z1-dev
\item libjpeg-turbo8-dev
\end{itemize}
On Ubuntu 16.04 it is possible to execute:
\begin{verbatim}
$ sudo apt-get install cmake libgtk2.0-dev pkg-config \
libavcodec-dev libavformat-dev libswscale-dev libjpeg8 \
libjpeg-turbo8-dbg lib32z1-dev libjpeg-turbo8-dev
\end{verbatim}
\item Then, it is necessary to install\textit{Opencv 2.4}
%
%
\item The code can be found in the B-Human branch called \textit{SPQRBallPerceptor} at the following address: \url{https://github.com/SPQRTeam/SPQRBallPerceptor}. Then, it is possible to install the Code as explained in the B-Human code release documentation.
\item Then, calibrate the Nao's camera matrix and joints to improve the detection and to not discard far away balls.
\end{itemize}

\section{Conclusions}
\label{sec:conclusions} 

In this technical report, we have described the SPQR Team approach
for ball perception in SPL competitions.
We have provided the details of our machine learning approach for
detecting the realistic black and white ball currently in use in the RoboCup
SPL. The detector is build on top of the B-Human framework and is based on a supervised approach. It consist of an LBP binary cascade classifier trained on our custom dataset. We provide a ready-to-use software module whose source code has been release and is freely available.

The presented approach allows for an accurate and precise ball detection, limiting dependencies on light conditions and colors. The SPQR Ball Perceptor be used not only for ball detection, but it is possible to extend the classifier also to different tasks; e.g. teammate recognition, field lines, landmarks detection and more.

As shown in the experimental validation, our approach can deal with the difficulties associated with variable
light conditions. The released code can be seen as a robust starting point for
the implementation of high-level functions to be
used in the future outdoors competitions, where external light changes, caused by variable whether,
clouds and other factors, can strongly affect the recognition tasks.

\section*{Acknowledgements}
Since 2013 SPQR Team uses the B-Human framework.
The work presented in this technical report has been developed
within the official B-Human code release 2016.
We would like to thank the B-Human Team for sharing their code.

\bibliographystyle{plain}
\bibliography{biblio}

\end{document}